\documentclass{article}
\usepackage{ijcai23}

\usepackage{times}
\usepackage{soul}
\usepackage{url}
\usepackage[hidelinks]{hyperref}
\usepackage[utf8]{inputenc}
\usepackage[small]{caption}
\usepackage{graphicx}
\usepackage{amsmath}
\usepackage{amsthm}
\usepackage{booktabs}
\usepackage{algorithm}
\usepackage{algorithmic}
\usepackage[switch]{lineno}

\usepackage{amsfonts,amssymb}
\usepackage{multirow}
\usepackage{physics}
\usepackage{color}

\title{RaSa: Relation and Sensitivity Aware Representation Learning \\for Text-based Person Search}

\author{
Yang Bai$^1$\and
Min Cao$^1$\thanks{Corresponding author}\and
Daming Gao$^1$\and
Ziqiang Cao$^1$\and
Chen Chen$^{2}$\and\\
Zhenfeng Fan$^{3}$\and
Liqiang Nie$^4$\And
Min Zhang$^{1,4}$
\affiliations
$^1$School of Computer Science and Technology, Soochow University\\
$^2$Institute of Automation, Chinese Academy of Sciences\\
$^3$Institute of Computing Technology, Chinese Academy of Sciences\\
$^4$Harbin Institute of Technology, Shenzhen
\emails
ybaibyougert@stu.suda.edu.cn,
mcao@suda.edu.cn
}

\begin{document}

\maketitle

\begin{abstract}
Text-based person search aims to retrieve the specified person images given a textual description. 
The key to tackling such a challenging task is to learn powerful multi-modal representations. 
Towards this, we propose a Relation and Sensitivity aware representation learning method (RaSa), including two novel tasks: Relation-Aware learning (RA) and Sensitivity-Aware learning (SA).
For one thing, existing methods cluster representations of all positive pairs without distinction and overlook the noise problem caused by the weak positive pairs where the text and the paired image have noise correspondences, thus leading to overfitting learning.
RA offsets the overfitting risk by introducing a novel positive relation detection task (\emph{i.e.}, learning to distinguish strong and weak positive pairs).
For another thing, learning invariant representation under data augmentation (\emph{i.e.}, being insensitive to some transformations) is a general practice for improving representation's robustness in existing methods. 
Beyond that, we encourage the representation to perceive the sensitive transformation by SA (\emph{i.e.}, learning to detect the replaced words), thus promoting the representation's robustness.
Experiments demonstrate that RaSa outperforms existing state-of-the-art methods by \textbf{6.94\%}, \textbf{4.45\%} and \textbf{15.35\%} in terms of Rank@1 on CUHK-PEDES, ICFG-PEDES and RSTPReid datasets, respectively.
Code is available at: \href{https://github.com/Flame-Chasers/RaSa}{https://github.com/Flame-Chasers/RaSa}.
\end{abstract}


\begin{figure}[ht]
\centering
\vspace{0.5cm}
\includegraphics[width=0.98\linewidth]{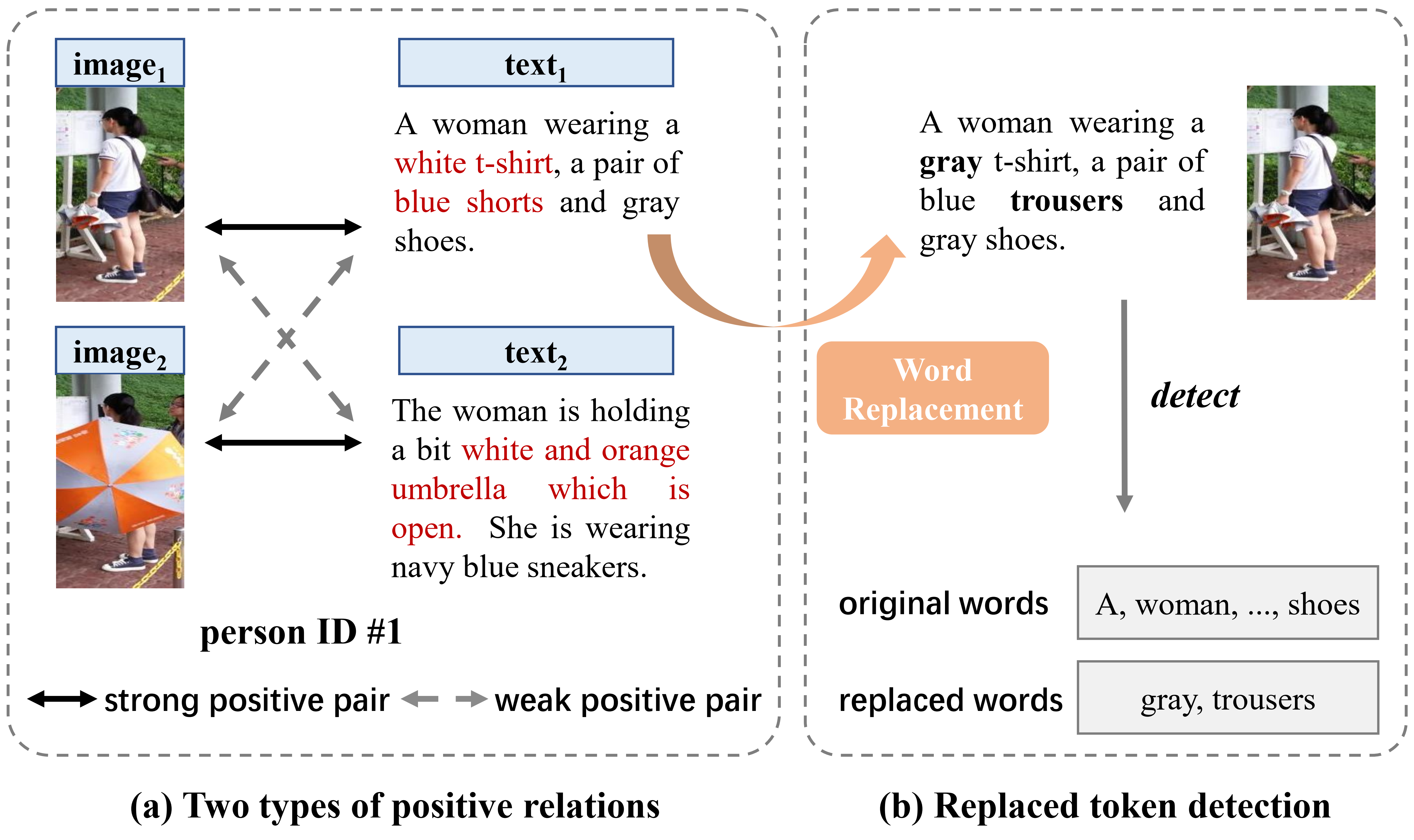}
\caption{Illustration of 
        (a) two types of positive relations for relation-aware learning, where the noise interference in the weak positive pairs is highlighted in red,
        (b) replaced token detection for sensitivity-aware learning, in which word replacement is used as the sensitive transformation and the replaced words are marked in bold.}
\label{fig1}
\end{figure}

\section{Introduction}
Text-based person search~\cite{li2017person,wang2021text} aims at retrieving the person images in a large-scale person image pool given a query of textual description about that person.
This task is related to person re-identification~\cite{ji2021context,wang2022pose} and text-image retrieval~\cite{ijcai2022-759,NEURIPS2021_50525975}, which have been very active research topics in recent years.
It, however, exhibits unique characteristics and challenges.
Compared to person re-identification with image queries, text-based person search with more accessible open-form text queries provides a more user-friendly searching procedure while embracing greater challenges due to the cross-modal search.
In addition, compared to general image-text retrieval, text-based person search focuses on cross-modal retrieval specific for the person with more fine-grained details, tending to larger intra-class variance as well as smaller inter-class variance, which toughly bottlenecks the retrieval performance.

Targeting learning powerful feature representation and achieving cross-modal alignment for text-based person search, 
researchers have developed a batch of technologies over the past few years~\cite{wu2021lapscore,shao2022learning}.
It has been proved that the model armed with reasonable tasks tends to learn better representation.  
In this paper, we propose a representation learning method, namely RaSa, with two novel tasks: relation-aware learning and sensitivity-aware learning for text-based person search.  

\textbf{Relation-aware learning.}
In existing methods \cite{han2021textreid,li2022learning}, the \emph{de facto} optimization objective is to bring image and text representations of the same identity (\emph{i.e.}, positive pairs) together and repel representations of different identities (\emph{i.e.}, negative pairs) away.
However, it tends to encounter the following issue.
Normally, a textual description is generated by annotating a particular single image in the text-based person search dataset. The text strongly matches the annotated image without a doubt, whereas it is not always well-aligned to other positive images of the same person at the semantic level due to intra-class variation in the image.
As shown in Figure~\ref{fig1} (a), the images and texts depict the same person, leading to a positive relation for each image-text pair.
However, there exist two different types of positive relations.
\textit{text$_1$} (\emph{resp.} \textit{text$_2$}) is the exact description of \textit{image$_1$} (\emph{resp.} \textit{image$_2$}), where they are completely matched and form a strong positive pair.
Nevertheless, \textit{image$_1$} and \textit{text$_2$} (\emph{resp.} \textit{image$_2$} and \textit{text$_1$}) constitute a weak positive pair with the noise interference.
For instance, ``white t-shirt" and ``blue shorts" in \textit{text$_1$} correspond to non-existent objects in \textit{image$_2$} due to the occlusion.
Existing methods endow the strong and weak positive pairs with equal weight in learning representations, regardless of the noise problem from the weak pairs, eventually leading to overfitting learning.


In order to mitigate the impacts of the noise interference from weak positive pairs, we propose a Relation-Aware learning (RA) task, which is composed of a probabilistic Image-Text Matching ($p$-ITM) task and a Positive Relation Detection (PRD) task.
$p$-ITM is a variant of the commonly-used ITM, aiming to distinguish negative and positive pairs with a probabilistic strong or weak positive inputting, while PRD is designed to explicitly makes a distinction between the strong and weak positive pairs.
Therein, $p$-ITM emphasizes the consistency between strong and weak positive pairs, whereas PRD highlights their difference and can be regarded as the regularization of $p$-ITM. 
The model armed with RA can not only learn valuable information from weak positive pairs by $p$-ITM but also alleviate noise interference from them by PRD, eventually reaching a trade-off.

\textbf{Sensitivity-aware learning.}
Learning invariant representations under a set of manually chosen transformations (also called \emph{insensitive} transformations in this context) is a general practice for improving the robustness of representation in the existing methods \cite{caron2020unsupervised,chen2021exploring}.
We recognize it but there is more.
Inspired by the recent success of equivariant contrastive learning~\cite{dangovski2022equivariant}, we explore the \emph{sensitive} transformation that would hurt performance when applied to learn transformation-invariant representations.
Rather than keeping invariance under insensitive transformation, we encourage the learned representations to have the ability to be aware of the sensitive transformation.

Towards this end, we propose a Sensitivity-Aware learning (SA) task.
We adopt the word replacement as the sensitive transformation and develop a Momentum-based Replaced Token Detection ($m$-RTD) pretext task to detect whether a token comes from the original textual description or the replacement, as shown in Figure~\ref{fig1} (b).
The closer the replaced word is to the original one (\emph{i.e.}, more confusing word), the more difficult this detection task is. 
When the model is trained to well solve such a detection task, it is expected to have the ability to learn better representation.
With these in mind, we use Masked Language Modeling (MLM) to perform the word replacement, which utilizes the image and the text contextual tokens to predict the masked tokens.
Furthermore, considering that the momentum model, a slow-moving average of the online model, can learn more stable representations than the current online model \cite{grill2020bootstrap} to generate more confusing words, we employ MLM from the momentum model to carry out the word replacement.
Overall, MLM and $m$-RTD together form a Sensitivity-Aware learning (SA), which offers powerful surrogate supervision for representation learning.

Our contributions can be summarized as follows:
\begin{itemize}
\item We differentiate between strong and weak positive image-text pairs in learning representation and propose a relation-aware learning task.
\item We pioneer the idea of learning representation under the sensitive transformation to the text-based person search and develop a sensitivity-aware learning task.
\item 
Extensive experiments demonstrate RaSa outperforms existing state-of-the-art methods by $6.94$\%, $4.45$\% and $15.35$\% in terms of Rank@1 metric on CUHK-PEDES, ICFG-PEDES and RSTPReid datasets, respectively.
\end{itemize}

\begin{figure*}[htp]
\setlength{\belowcaptionskip}{0.2cm}
\centering
\includegraphics[width=0.95\linewidth]{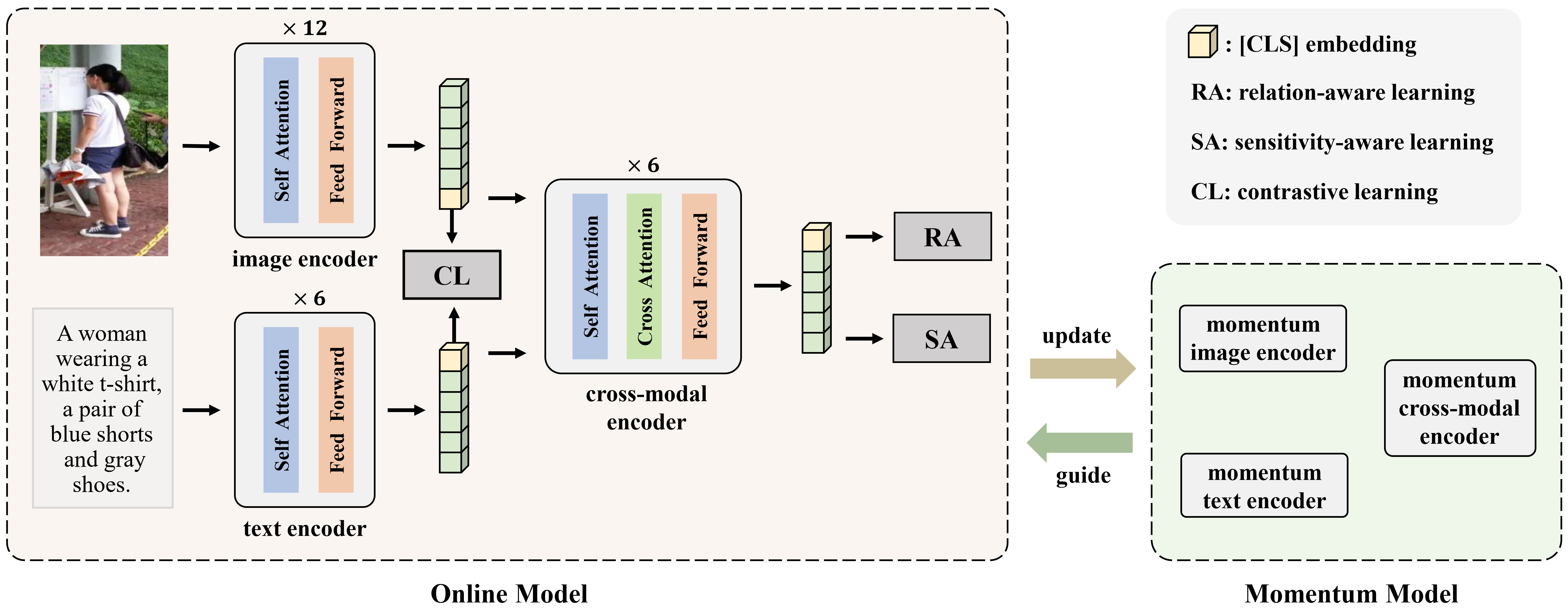}
\caption{Model architecture of RaSa. It consists of an image encoder, a text encoder and a cross-modal encoder.
An intra- and cross-modal CL task is attached after the unimodal encoders for unimodal representation learning. RA and SA tasks are tied after the cross-modal encoders for multi-modal representation learning. The momentum model (a slow-moving of the online model) is used to guide the online model to learn better representations.}
\label{fig2}
\end{figure*}

\section{Related Work}
\subsubsection{Text-based Person Search}
\citeauthor{li2017person}~\shortcite{li2017person} first introduce the text-based person search task and publish a challenging dataset CUHK-PEDES. Following this, a series of methods are proposed to solve this task.
Part of methods \cite{zheng2020hierarchical,wang2021text} focus on designing a reasonable cross-modal alignment strategy, 
while others~\cite{zhang2018deep,shao2022learning} concentrate on learning powerful feature representation. 
For cross-modal alignment,
it begins with global alignment \cite{zheng2020dual} or local correspondences (\emph{e.g.}, patch-word or region-phrase correspondences) \cite{chen2022tipcb,niu2020improving}, and evolves into self-adaptively learning semantic alignment across different granularity~\cite{li2022learning,gao2021contextual}. 
Beyond that, some works \cite{wang2020vitaa,zhu2021dssl} utilize external technologies (\emph{e.g.}, human segmentation, pose estimation or attributes prediction) to assist with the cross-modal alignment.
For representation learning,
\citeauthor{wu2021lapscore}~\shortcite{wu2021lapscore} propose two color-related tasks based on the observation that color plays a key role in text-based person search.
\citeauthor{zeng2021relation}~\shortcite{zeng2021relation} develop three auxiliary reasoning tasks with gender classification, appearance similarity and image-to-text generation.
\citeauthor{ding2021semantically}~\shortcite{ding2021semantically} firstly notice the noise interference from weak positive pairs and propose to keep the difference between strong and weak positive pairs by manually assigning different margins in the triplet loss.
More recently, some works~\cite{han2021textreid,shu2022see,yan2022clip} resort to vision-language pretraining models to learn better representations.
In this paper, we design two novel tasks: RA and SA.
RA detects the type of the positive pair to weaken noise from weak positive pairs, differently from the method~\cite{ding2021semantically} with the sophisticated trick.
SA focuses on representation learning by detecting sensitive transformation, which is under-explored in the previous methods.

\subsubsection{Equivariant Contrastive Learning}
Different from contrastive learning~\cite{he2020momentum} that aims to learn transformation-insensitive representations, equivariant contrastive learning~\cite{dangovski2022equivariant} is recently proposed by additionally encouraging the learned representations to have the ability to be aware of sensitive transformations.
Mathematically, the notions of insensitivity and sensitivity can be inductively summarized as: $f(T(x)) = T'(f(x))$ where $T$ denotes a group of transformations of an input instance $x$, and $f$ is an encoder to compute the representation of $x$.
When $T'$ is the identity transformation, it can be said that $f$ is trained to be insensitive to $T$; otherwise, $f$ is sensitive to $T$.
Equivariant contrastive learning has shown its successful application in the fields of computer vision (CV)~\cite{dangovski2022equivariant} and natural language processing (NLP)~\cite{chuang2022diffcse}, which inspires us to explore sensitive transformations for learning high-quality representations in the cross-modal retrieval task.
In this paper, we develop a sensitivity-aware learning with MLM-based word replacement as the sensitive transformation to encourage the model to perceive the replaced words, thus obtaining more informative and discriminative representations.

\section{Method}
In this section, we take ALBEF~\cite{NEURIPS2021_50525975} as the backbone\footnote{Experiments on more backbones are shown in Appendix~\ref{backbones}.} and elaborate on the proposed method RaSa by introducing the modal architecture in Section~\ref{MA} and the optimization objectives involving the proposed RA and SA tasks in Section~\ref{OB}.

\subsection{Model Architecture}
\label{MA}
As illustrated in Figure~\ref{fig2}, the proposed RaSa consists of two unimodal encoders and a cross-modal encoder.
We adopt $12$-layer and $6$-layer transformer blocks for the image and text encoders, respectively.
The cross-modal encoder comprises $6$-layer transformer blocks, where a cross-attention module is added after the self-attention module in each block.
Considering that the textual description usually covers a part of the information in the corresponding image, we employ a text-guided asymmetric cross-attention module in the cross-modal encoder, \emph{i.e.}, using the textual representation as query and the visual one as key and value.
Simultaneously, we maintain a momentum version of the online model via Exponential Moving Average (EMA).
Specifically, EMA is formulated as $\hat{\theta} = m\hat{\theta} + (1-m)\theta$, where $\hat{\theta}$ and $\theta$ are the parameters of the momentum and online models, respectively, and $m \in [0, 1]$ is a momentum coefficient.
The momentum model presents a delayed and more stable version of the online model and is used to guide the online model to learn better representations. 

Given an image-text pair $(I, T)$, we first feed the image $I$ into the image encoder to obtain a sequence of visual representations $\{v_{cls}, v_1,\cdots,v_M\}$ with $v_{cls}$ being the global visual representation and $v_i$ $(i=1,\cdots,M)$ being the patch representation.
Similarly, we obtain a sequence of textual representations $\{t_{cls}, t_1,\cdots,t_N\}$ by feeding the text $T$ into the text encoder, where $t_{cls}$ is the global textual representation and $t_i$ $(i=1,\cdots,N)$ is the token representation.
The visual and textual representations are then fed to the cross-modal encoder to obtain a sequence of multi-modal representations $\{f_{cls}, f_1,\cdots,f_N\}$, where $f_{cls}$ denotes the joint representation of $I$ and $T$, and $f_i$ $(i=1,\cdots,N)$ can be regarded as the joint representation of the image $I$ and the $i$-th token in the text $T$. 
Simultaneously, the momentum model is employed to obtain a sequence of momentum representations.

\subsection{Optimization Objectives}
\label{OB}
\subsubsection{Relation-aware Learning}
The vanilla widely-used ITM predicts whether an inputted image-text pair is positive or negative, defined as:
\begin{equation}
    L_{itm} = \mathbb{E}_{p(I, T)}{\mathcal{H}}(y^{itm}, \phi^{itm}(I, T)), \label{equ1}
\end{equation}
where $\mathcal{H}$ represents a cross-entropy function, $y^{itm}$ is a $2$-dimension one-hot vector representing the ground-truth label (\emph{i.e.}, $\left[0, 1\right]^\top$ for the positive pair, and $\left[1, 0\right]^\top$ for the negative pair), and $\phi^{itm}(I, T)$ is the predicted matching probability of the pair that is computed by feeding $f_{cls}$ into a binary classifier, a fully-connected layer followed by a softmax function.

However, it is unreasonable to directly adopt the vanilla ITM in text-based person search.
On the one hand, there exists noise interference from weak positive pairs, which would hamper the representation learning.
On the other hand, the weak positive pairs contain certain valuable alignment information that can facilitate representation learning.
As a result, to reach a balance, we retain a proportion of weak positive pairs in ITM by introducing the probabilistic inputting.
Specifically, we input the weak positive pair with a small probability of $p^w$ and the strong positive pair with a probability of $1-p^w$.
To distinguish with the vanilla ITM, we denote the proposed probabilistic ITM as $p$-ITM.

Furthermore, we continue to alleviate the noise effect of the weak pairs.
We propose a Positive Relation Detection (PRD) pretext task to detect the type of the positive pair (\emph{i.e.}, strong or weak), which is formulated as:
\begin{equation}
    L_{prd} = \mathbb{E}_{p({I}, {T^p})}{\mathcal{H}}(y^{prd}, \phi^{prd}({I}, {T^p})), \label{equ2}
\end{equation}
where $({I}, {T^p})$ denotes a positive pair, $y^{prd}$ is the ground truth label (\emph{i.e.}, $\left[1, 0\right]^\top$ for the strong positive pair and $\left[0, 1\right]^\top$ for the weak pair), and $\phi^{prd}({I}, {T^p})$ is the predicted probability of the pair which is computed by appending a binary classifier to the joint representation $f_{cls}$ of the pair.

Taken together, we define the Relation-Aware learning (RA) task as:
\begin{equation}
    L_{ra} = L_{p\raisebox{0mm}{-}itm} + \lambda_1L_{prd}, \label{equ3}
\end{equation}
where the weight $\lambda_1$ is a hyper-parameter.

During the process of the optimization, $p$-ITM focuses on the consistency between strong and weak positive pairs, while PRD highlights their difference.
In essence, PRD plays a role of a regularized compensation for $p$-ITM. 
As a whole, RA achieves a trade-off between the benefits of the weak pair and the risk of its side effects.

\subsubsection{Sensitivity-aware Learning}
Learning invariant representations under the \emph{insensitive} transformation of data is a common way to enhance the robustness of the learned representations.
We go beyond it and propose to learn representations that are aware of the \emph{sensitive} transformation.
Specifically, we adopt the MLM-based word replacement as the sensitive transformation and propose a Momentum-based Replaced Token Detection ($m$-RTD) pretext task to detect (\emph{i.e.}, being aware of) the replacement.

\begin{figure}[t]
\centering
\includegraphics[width=0.95\linewidth]{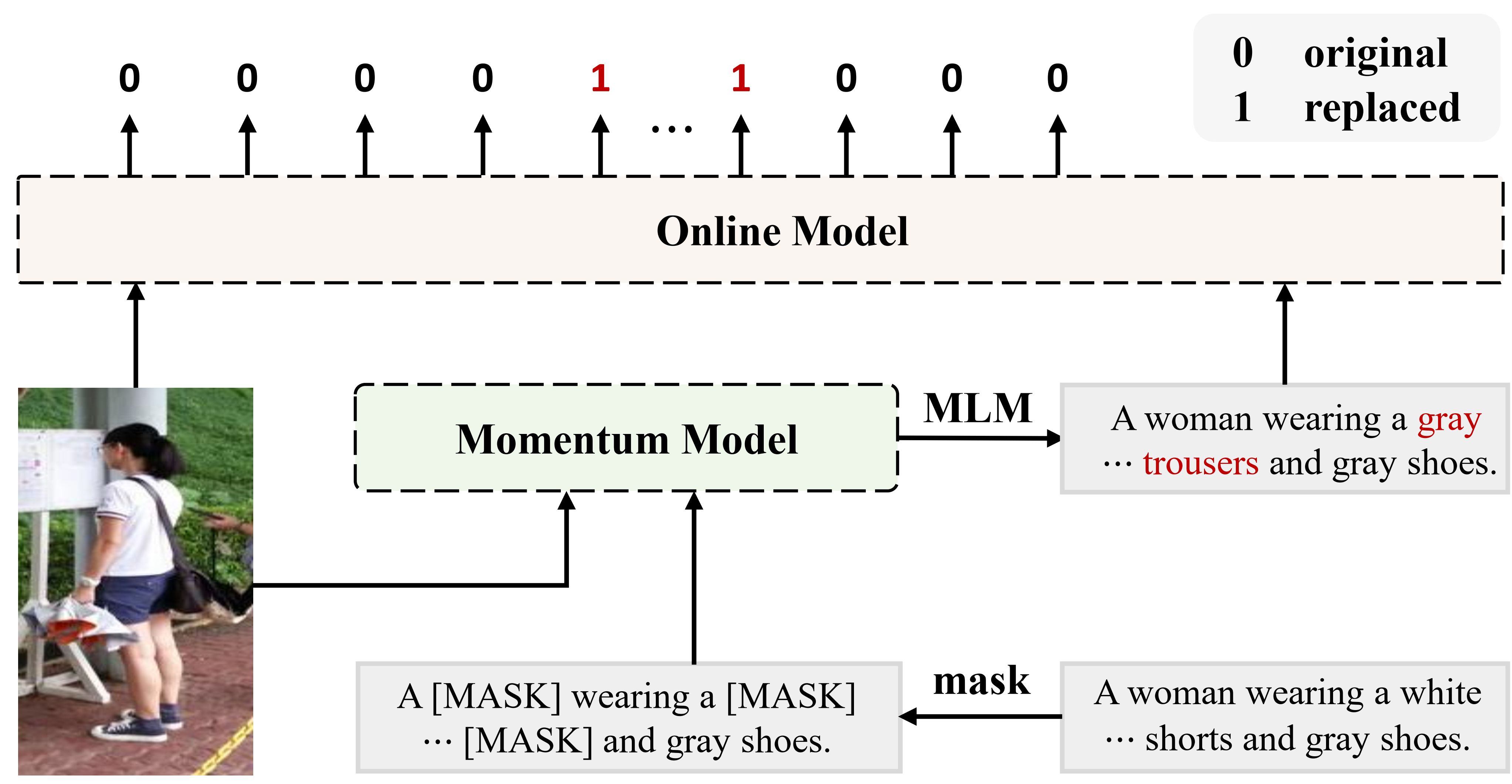} 
\caption{Illustration of $m$-RTD. It aims to detect whether a token is from the original textual description or the replacement with the aid of the information of the contextual tokens and the paired image. The text with word replacement is obtained by the result of the Masked Language Modeling (MLM) from the momentum model.}
\label{fig3}
\end{figure}

Given a strong positive pair $(I, T^s)$, MLM loss is formulated as:
\begin{equation}
    L_{mlm} = \mathbb{E}_{p(I, T^{msk})}{\mathcal{H}}(y^{mlm}, \phi^{mlm}(I, T^{msk})), \label{equ4}
\end{equation}
where $T^{msk}$ is a masked text in which each token in the input text $T^s$ is randomly masked with a probability of $p^{m}$, $y^{mlm}$ is a one-hot vector denoting the ground truth of the masked token and $\phi^{mlm}(I, T^{msk})$ is the predicted probability for the masked token based on the information of the contextual text $T^{msk}$ and the paired image $I$. 

We use the result of MLM from the momentum model as the word replacement, denoted as $T^{rep}$.
The momentum model is a slow-moving of the online model and can learn more stable representations. Therefore, the momentum model is expected to generate more confusing tokens.
As $m$-RTD detects such challenging tokens well, the model is motivated to learn more informative representations to distinguish the tiny differences.
Remarkably, besides serving as a generator for the word replacement, MLM also plays a role of token-level optimization, promoting fine-grained representation learning.

Next, $m$-RTD performs a detection of the MLM-based token replacement. Specifically, the pair $(I, T^{rep})$ is inputted to the model to obtain a sequence of multi-modal representations $\{f_{cls}, f_1, ..., f_N\}$, and a binary classifier works on $\{f_1, ..., f_N\}$ to predict whether the $i$-th token is replaced or not.
$m$-RTD minimizes a cross-entropy loss:
\begin{equation}
    L_{m\raisebox{0mm}{-}rtd} = \mathbb{E}_{p(I, T^{rep})}{\mathcal{H}}(y^{m\raisebox{0mm}{-}rtd}, \phi^{m\raisebox{0mm}{-}rtd}(I, T^{rep})), \label{equ5}
\end{equation}
where $y^{m\raisebox{0mm}{-}rtd}$ is a one-hot vector denoting the ground truth of the replaced token and $\phi^{m\raisebox{0mm}{-}rtd}(I, T^{rep})$ is the predicted replacement probability.
We illustrate the pipeline of $m$-RTD in Figure \ref{fig3} for clarity.

Overall, Sensitivity-Aware learning (SA) loss is defined as:
\begin{equation}
    L_{sa} = L_{mlm} + \lambda_2L_{m\raisebox{0mm}{-}rtd}, \label{equ6}
\end{equation}
where the weight $\lambda_2$ is a hyper-parameter.

In conclusion, RA works on the global representation $f_{cls}$ and mainly focuses on the correlation between the image and text, which can be regarded as a coarse-grained optimization.
As a complement, SA acts on the token representations $\{f_1, ..., f_N\}$ and pays more attention to the interaction between the image and textual tokens, exhibiting a fine-grained optimization.
The two complementary tasks effectively facilitate representation learning.

\subsubsection{Contrastive Learning}
The proposed RA and SA are directly applied on the multi-modal representations from the cross-modal encoder.
Furthermore, we introduce an intermediate Contrastive Learning task (CL) on the representations from the unimodal encoders, so as to make the subsequent cross-modal fusion easier to perform multi-modal representation learning.

Given an image-text pair $(I, T)$, we feed it into the unimodal encoders and obtain the global visual and textual representations $v_{cls}$ and $t_{cls}$.
Then a linear layer is applied to project them to lower-dimensional representations $v'_{cls}$ and $t'_{cls}$.
Meanwhile, we obtain the output of momentum unimodal encoders, denoted as $\hat{v}'_{cls}$ and $\hat{t}'_{cls}$.
We maintain an image queue $\hat{Q}_v$ and a text queue $\hat{Q}_t$ to store the recent $R$ projected representations $\hat{v}'_{cls}$ and $\hat{t}'_{cls}$, similarly to MoCo~\cite{he2020momentum}. 
The introduction of the queues implicitly enlarges the batch size, and a larger batch will provide more negative samples, thereby facilitating representation learning.

In CL, the general form of InfoNCE loss is formulated as:
\begin{small}
\begin{equation}
    L_{nce}(x, x_+, Q) = -\mathbb{E}_{p(x, x_+)} [\log\frac{\exp(s(x, x_+)/\tau)}{\sum\limits_{x_i \in Q}\exp(s(x, x_i)/\tau)}], \label{equ7}
\end{equation}
\end{small}
where $\tau$ is a learnable temperature parameter, $Q$ denotes a maintained queue, and $s(x, x_+)=x^{\rm T}x_+/\lVert x\rVert\lVert x_+\rVert$ measures the cosine similarity between $x$ and $x_+$. 

Beyond the widely-used cross-modal image-text contrastive learning (ITC)~\cite{NEURIPS2021_50525975,radford2021learning}, denoted as:
\begin{equation}
    L_{itc} = [L_{nce}(v'_{cls}, \hat{t}'_{cls}, \hat{Q}_t) + L_{nce}(t'_{cls}, \hat{v}'_{cls}, \hat{Q}_v)]\ /\ 2, \label{equ8}
\end{equation}
we additionally explore the intra-modal contrastive learning (IMC).
The representations of the same person are supposed to stay closer than those of different persons within each modality. IMC loss is formulated as:
\begin{equation}
    L_{imc} = [L_{nce}(v'_{cls}, \hat{v}'_{cls}, \hat{Q}_v) + L_{nce}(t'_{cls}, \hat{t}'_{cls}, \hat{Q}_t)]\ /\ 2. \label{equ9}
\end{equation}

Taken together, we define the overall loss for CL as:
\begin{equation}
    L_{cl} = (L_{itc} + L_{imc})\ /\ 2. \label{equ10}
\end{equation}



\subsubsection{Joint Learning}
Overall, we formulate the joint optimization objective as:
\begin{equation}
    L = L_{ra} + L_{sa} + \lambda_3L_{cl}, \label{equ11}
\end{equation}
where $\lambda_3$ is a hyper-parameter.

During inference, given a query text and a large-scale image pool, we use the predicted matching probability from $p$-ITM to rank all images. 
Considering the inefficiency of the cross-modal encoder with quadratic interaction operation, we refer to ALBEF~\cite{NEURIPS2021_50525975} and exclude a large number of irrelevant image candidates prior to the cross-modal encoder, thereby speeding up the inference.
Specifically, we first calculate each pair's similarity $s(t_{cls}, v_{cls})$ via the unimodal encoders, and then select the first $128$ images with the highest similarities to send them to the cross-modal encoder and compute the $p$-ITM matching probabilities for ranking.

\section{Experiments}

We conduct experiments on three text-based person search datasets: CUHK-PEDES~\cite{li2017person}, ICFG-PEDES~\cite{ding2021semantically} and RSTPReid~\cite{zhu2021dssl}.
\emph{The introduction of each dataset and the implementation details of the proposed method are shown in Appendix~\ref{datasets} and \ref{Implementation Details}, respectively.}

\subsection{Evaluation Protocol}
We adopt the widely-used Rank@K (R@K for short, K=$1,5,10$) metric to evaluate the performance of the proposed method. 
Specifically, given a query text, we rank all the test images via the similarity with the text and the search is deemed to be successful if top-K images contain any corresponding identity. R@K is the percentage of successful searches. 
We also adopt the mean average precision (mAP) as a complementary metric.

\begin{table}[t]
\small
\centering
\tabcolsep=2pt
\renewcommand\arraystretch{1.1}
\begin{tabular}{c|l|cccc}
\hline
                                               & Method    & R@1  & R@5  & R@10   & mAP \\
\hline
\multirow{11}{*}{\rotatebox{90}{w/o VLP}}      & GNA-RNN \cite{li2017person}      & 19.05     & -      & 53.64      & -      \\
                                               & Dual Path \cite{zheng2020dual}   & 44.40     & 66.26      & 75.07      & -      \\
                                               & CMPM/C \cite{zhang2018deep}      & 49.37     & 71.69      & 79.27      & -      \\
                                               & ViTAA \cite{wang2020vitaa}       & 55.97     & 75.84      & 83.52      & -      \\
                                               & DSSL \cite{zhu2021dssl}          & 59.98     & 80.41      & 87.56      & -      \\
                                               & MGEL \cite{wang2021text}         & 60.27     & 80.01      & 86.74      & -      \\
                                               & ACSA \cite{ji2022asymmetric}     & 63.56     & 81.40      & 87.70      & -  \\
                                               & SAF \cite{li2022learning}        & 64.13     & 82.62      & 88.40      & 58.61  \\
                                               & TIPCB \cite{chen2022tipcb}       & 64.26     & 83.19      & 89.10      & -      \\
                                               & CAIBC \cite{wang2022caibc}       & 64.43     & 82.87      & 88.37      & -      \\
                                               & $\rm C_2A_2$ \cite{niu2022cross} & 64.82     & 83.54      & 89.77      & -  \\
                                               & LGUR \cite{shao2022learning}     & 65.25     & 83.12      & 89.00      & -  \\
\hline\hline
\multirow{5}{*}{\rotatebox{90}{w/ VLP}}        & PSLD \cite{han2021textreid}      & 64.08     & 81.73      & 88.19      & 60.08  \\
                                               & IVT \cite{shu2022see}            & 65.59     & 83.11      & 89.21      & -  \\
                                               & CFine \cite{yan2022clip}         & 69.57     & 85.93      & 91.15      & -  \\
\cline{2-6}
                                               & ALBEF(backbone) \cite{NEURIPS2021_50525975}& 60.28     & 79.52      & 86.34      & 56.67           \\
                                               & \textbf{RaSa (Ours)}        & \textbf{76.51}    & \textbf{90.29}    & \textbf{94.25}    & \textbf{69.38} \\
\hline
\end{tabular}
\caption{Comparison with other methods on CUHK-PEDES. VLP denotes vision-language pretraining. For a fair comparison, all reported results come from the methods without re-ranking.}
\label{table1}
\end{table}

\subsection{Backbones}
\label{Baselines}


Most text-based person search methods~\cite{li2022learning,shao2022learning} rely on two feature extractors pre-trained on unaligned images and texts separately, such as ResNet~\cite{he2016deep} or ViT~\cite{dosovitskiy2020image} for the visual extractor, Bi-LSTM~\cite{hochreiter1997long} or BERT~\cite{devlin2018bert} for the textual extractor. 
Recently, some works~\cite{shu2022see,yan2022clip} have applied vision-language pretraining (VLP) to text-based person search and obtained impressive results. 
Following this, we adopt VLP models as the backbone.

The proposed RaSa can be plugged into various backbones. 
To adequately verify the effectiveness, we conduct RaSa on three VLP models: ALBEFF~\cite{NEURIPS2021_50525975}, TCL~\cite{yang2022vision} and CLIP~\cite{radford2021learning}. 
We use ALBEF as the backbone by default in the following experiments, which is pre-trained on $14$M image-text pairs and adopts ITC and ITM tasks for image-text retrieval.
\emph{The details and experiments on TCL and CLIP are shown in Appendix~\ref{backbones}.}

\begin{table}[t]
\small
\centering
\tabcolsep=2pt
\renewcommand\arraystretch{1.1}
\begin{tabular}{c|l|cccc}
\hline
                                               & Method    & R@1  & R@5  & R@10   & mAP \\
\hline
\multirow{8}{*}{\rotatebox{90}{w/o VLP}}       & Dual Path \cite{zheng2020dual}   & 38.99     & 59.44     & 68.41      & -      \\
                                               & CMPM/C \cite{zhang2018deep}      & 43.51     & 65.44     & 74.26      & -      \\
                                               & ViTAA \cite{wang2020vitaa}       & 50.98     & 68.79     & 75.78      & -      \\
                                               & SSAN \cite{ding2021semantically} & 54.23     & 72.63     & 79.53      & -      \\
                                               & SAF \cite{li2022learning}        & 54.86     & 72.13     & 79.13      & 32.76    \\
                                               & TIPCB \cite{chen2022tipcb}       & 54.96     & 74.72     & 81.89      & -      \\
                                               & SRCF \cite{suo2022simple}        & 57.18     & 75.01     & 81.49      & -      \\
                                               & LGUR \cite{shao2022learning}     & 59.02     & 75.32     & 81.56      & -  \\
\hline\hline
\multirow{4}{*}{\rotatebox{90}{w/ VLP}}        
                                               & IVT \cite{shu2022see}            & 56.04     & 73.60     & 80.22      & -  \\
                                               & CFine \cite{yan2022clip}         & 60.83     & 76.55     & 82.42      & -  \\
\cline{2-6}
                                               & ALBEF(backbone) \cite{NEURIPS2021_50525975}& 34.46     & 52.32      & 60.40      & 19.62 \\
                                               & \textbf{RaSa (Ours)}        & \textbf{65.28} & \textbf{80.40} & \textbf{85.12} & \textbf{41.29} \\
\hline
\end{tabular}
\caption{Comparison with other methods on ICFG-PEDES.}
\label{table2}
\end{table}

\begin{table}[t]
\small
\centering
\tabcolsep=2pt
\renewcommand\arraystretch{1.1}
\begin{tabular}{c|l|cccc}
\hline
                                               & Method    & R@1  & R@5  & R@10   & mAP \\
\hline
\multirow{6}{*}{\rotatebox{90}{w/o VLP}}       & DSSL \cite{zhu2021dssl}          & 32.43     & 55.08      & 63.19      & -      \\
                                               & SSAN \cite{ding2021semantically} & 43.50     & 67.80      & 77.15      & -      \\
                                               & SAF \cite{li2022learning}        & 44.05     & 67.30      & 76.25      & 36.81    \\
                                               & CAIBC \cite{wang2022caibc}       & 47.35     & 69.55      & 79.00      & -      \\
                                               & ACSA \cite{ji2022asymmetric}     & 48.40     & 71.85      & 81.45      & -  \\
                                               & $\rm C_2A_2$ \cite{niu2022cross} & 51.55     & 76.75      & 85.15      & -  \\
\hline\hline
\multirow{4}{*}{\rotatebox{90}{w/ VLP}}        & IVT \cite{shu2022see}            & 46.70     & 70.00     & 78.80      & -  \\
                                               & CFine \cite{yan2022clip}         & 50.55     & 72.50     & 81.60      & -  \\
\cline{2-6}
                                               & ALBEF(backbone) \cite{NEURIPS2021_50525975}& 50.10     & 73.70     & 82.10      & 41.73           \\
                                               & \textbf{RaSa (Ours)}        & \textbf{66.90} & \textbf{86.50} & \textbf{91.35} & \textbf{52.31} \\
\hline
\end{tabular}
\caption{Comparison with other methods on RSTPReid.}
\label{table3}
\end{table}

\subsection{Comparison with State-of-the-art Methods}
We compare the proposed RaSa with the existing text-based person search methods on CUHK-PEDES, ICFG-PEDES and RSTPReid, as shown in Table \ref{table1}, \ref{table2} and \ref{table3}, respectively.    
RaSa achieves the highest performance in terms of all metrics, outperforming existing state-of-the-art methods by a large margin. 
Specifically, compared with the current best-performing method CFine~\cite{yan2022clip}, RaSa gains a significant R@1 improvement of $6.94$\%, $4.45$\% and $15.35$\% on the three datasets, respectively.
The comparison clearly demonstrates the effectiveness of RaSa in text-based person search.

\subsection{Ablation Study}

We analyze the effectiveness and contribution of each optimization objective in RaSa by conducting a series of ablation experiments on CUHK-PEDES, as shown in Table~\ref{table4}.

\subsubsection{Effectiveness of Optimization Objectives}
RaSa consists of three optimization objectives. CL provides an explicit alignment before the cross-modal fusion. RA implements the deep fusion by the cross-modal encoder with an alleviation of noise interference. And SA encourages the learned representations to be sensitive to the MLM-based token replacement.

\begin{figure*}[ht]
\setlength{\belowcaptionskip}{0.35cm}
\centering
\includegraphics[width=0.95\textwidth, height=0.17\textwidth]{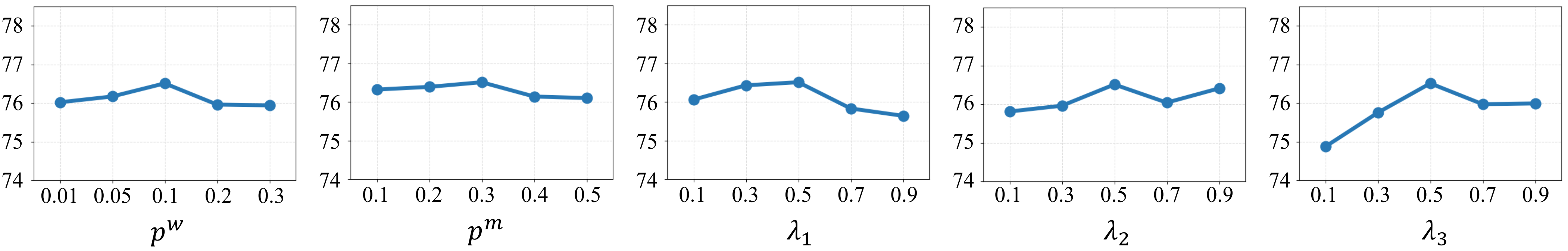} 
\caption{The impact of the hyper-parameters at R@1 on CUHK-PEDES. $p^w$ denotes the probability of inputting weak positive pairs in RA. $p^m$ means the masking ratio of the tokens in a text in SA. $\lambda_1$, $\lambda_2$ and $\lambda_3$ are the loss weights.}
\label{fig4}
\end{figure*}

We can see from Table~\ref{table4},
(1) RaSa with a single CL achieves a modest performance of $61.35$\% and $59.44$\% in terms of R@1 and mAP, respectively. On account of the modality gap between the image and text and the fine-grained intra-class variation, CL contributes a coarse alignment with a lack of deep interaction across modalities, which is not enough to handle such a challenging retrieval task.
(2) When adding RA($p$-ITM + PRD), the performance has a remarkable improvement of $12.85$\% at R@1 and $8.67$\% at mAP, effectively demonstrating that deep cross-modal fusion with RA is extraordinarily significant to text-based person search.
And (3) with the aid of SA(MLM + $m$-RTD), RaSa achieves the best performance of $76.51$\% at R@1 and $69.38$\% at mAP. SA utilizes the visual information and the contextual token information of the corresponding text to detect whether a token has been replaced or not. In order to handle such a challenging detection task, the learned representations are encouraged to be powerful enough to distinguish the tiny difference between the original token and the replaced one.

\begin{table}[tp]
\small
\renewcommand\arraystretch{1.2}
\setlength{\tabcolsep}{5pt}
\centering
\begin{tabular}{c|l|cccc} 
\hline
Module              & Setting         & R@1            & R@5            & R@10           & mAP             \\ 
\hline
CL                  & ITC + IMC         & 61.35          & 80.44          & 86.91          & 59.44           \\ 
\hline
\multirow{5}{*}{+RA} & ITM            & 71.29          & 86.70          & 91.46          & 67.82           \\
                    & $s$-ITM          & 73.52          & 88.71          & 92.98          & 66.74           \\
                    & $p$-ITM          & 72.58          & 87.98          & 92.51          & 68.29           \\
                    & ITM + PRD        & 73.03          & 87.75          & 92.45          & 68.45           \\
                    & $p$-ITM + PRD      & 74.20          & 89.02          & 92.95          & 68.11           \\ 
\hline
\multirow{4}{*}{++SA} & MLM         & 74.81          & 89.85          & 93.66          & 68.32           \\
                    & MLM + $f$-RTD & 75.13          & 89.93          & 93.47          & 69.17           \\
                    & MLM + $o$-RTD & 75.99          & 90.21          & 94.09          & 69.35           \\
                    & MLM + $m$-RTD & 76.51          & 90.29          & 94.25          & 69.38  \\
\hline
\end{tabular}
\caption{Comparison of RaSa with different settings on CUHK-PEDES. ITM learns from all positive pairs without a probabilistic inputting. $s$-ITM learns from only strong positive pairs and discards all weak positive pairs. $p$-ITM uses a probabilistic inputting of strong and weak positive pairs. $f$-RTD adopts DistilBERT~\protect\cite{sanh2019distilbert} as a fixed generator to produce the replaced tokens. $o$-RTD uses the online model as the generator, while $m$-RTD is based on the momentum model.}
\label{table4}
\end{table}

\subsubsection{Analysis of RA}
RA contains $p$-ITM and PRD, where the former focuses on the consistency between the strong and weak positive pairs, while the latter highlights their difference, serving as a regularization of $p$-ITM. 

The vanilla ITM learns from all positive pairs without the probabilistic inputting.
However, there exists too much noise interference from weak positive pairs.
Intuitively, we can discard all weak positives to get rid of the noise. 
$s$-ITM only uses the strong positive pairs and gains a boost of $2.23$\% at R@1 compared to the vanilla ITM.
Nevertheless, such a straightforward way ignores the weak supervision from the weak positives which is also beneficial to representation learning. 
To reach a trade-off between the benefits of the weak supervision and the risk of side effects, $p$-ITM resorts to the probabilistic inputting and retains a small proportion of the weak positives.
Compared with the vanilla ITM and $s$-ITM, $p$-ITM achieves an intermediate performance.
Not surprisingly at all, the more noise there exists, the more it affects the retrieval performance.
In order to alleviate the impact of the noise, we further propose PRD to perform an explicit distinction between the strong and weak positives, which serve as a regularization for $p$-ITM.
Significantly, no matter whether adding PRD to the vanilla ITM or $p$-ITM, PRD can obtain consistent performance improvement, which powerfully demonstrates its effectiveness.

\subsubsection{Analysis of SA}
SA includes MLM and $m$-RTD. MLM not only plays the role of generating the text with word replacement but also performs a token-level optimization. $m$-RTD detects the replaced tokens by virtue of the visual information and the contextual token information.

Based on CL and RA, adding a single MLM without the replacement detection task brings a slight boost of $0.61$\% at R@1.
Furthermore, we introduce the detection task and use the momentum model as the generator to produce the replaced tokens.
In order to adequately investigate the effectiveness of the generator, we compare three different variants.
(1) Following DiffCSE~\cite{chuang2022diffcse}, we use DistilBERT~\cite{sanh2019distilbert} as a fixed generator for the word replacement, which is denoted as $f$-RTD. 
From Table~\ref{table4}, RaSa with $f$-RTD gains a modest performance of $75.13$\% at R@1. 
We argue that the generated tokens from a fixed generator can be easily detected as the training advances and thus provides a limited effect on learning representation. 
(2) $o$-RTD adopts the online model as the generator. 
RaSa with $o$-RTD achieves a better performance of $75.99\%$ at R@1. 
Compared with $f$-RTD, $o$-RTD resorts to a dynamic generator which is optimized constantly during the whole training process and can produce more confusing tokens with the proceeding of the model's training, effectively increasing the difficulty of replaced tokens detection and facilitating representation learning.
And (3) $m$-RTD adopts the momentum model as the generator and reaches the best performance of $76.51$\% at R@1. The momentum model is a slow-moving of the online model and can obtain more stable representations. 
As the training goes ahead, the momentum model iteratively bootstraps MLM to generate more challenging tokens for detection, which encourages the learned representations to be powerful enough to distinguish the tiny difference and substantially improve results.

\subsubsection{Hyper-parameters}
In Section~\ref{OB}, we use the inputting probability $p^{w}$ to retain a small proportion of weak positive pairs to alleviate the noise, the masking ratio $p^{m}$ to randomly mask tokens to perform the replaced token detection, and the loss weights $\lambda_1$, $\lambda_2$, $\lambda_3$ to make a trade-off. 
We show how these hyper-parameters impact the performance of RaSa in Figure \ref{fig4}.
(1) The best result is achieved at $p^{w}=0.1$. The inputting probability $p^{w}$ in RA is introduced to seek a balance between the useful information and the noise from weak positives. A larger $p^{w}$ may introduce too much noise, while a smaller $p^{w}$ hinders the model from making full use of the useful information. 
(2) RaSa performs best at $p^{m}=0.3$. A larger $p^{m}$ brings more perturbations to the text, making the detection task too difficult to be carried out. In contrast, when $p^{m}$ goes smaller, SA will contribute less to representation learning. 
And (3) for the loss weights $\lambda_1$, $\lambda_2$ and $\lambda_3$, they present an overall trend of first increasing and then decreasing. Empirical results show that RaSa performs best when they are set as $0.5$.

\subsection{Extended Experiments and Visualization}


To go a step further and validate the effectiveness of RaSa, we perform extended experiments on two coarse-grained image-text retrieval datasets (Flickr30K~\cite{plummer2015flickr30k} and COCO~\cite{lin2014microsoft}), as well as two fine-grained datasets (CUB~\cite{reed2016learning} and Flowers~\cite{reed2016learning}). 
\emph{The experimental results are shown in Appendix~\ref{Extended Experiments}.}
Besides, we conduct a series of domain generalization experiments following LGUR~\cite{shao2022learning} in Appendix~\ref{Extended Experiments} to verify the generalization ability of RaSa. These results clearly demonstrate the effectiveness and the generalization ability of RaSa. 

For a qualitative analysis, we also present the retrieval visualization in Appendix~\ref{visualization}, vividly showing the excellent retrieval ability of RaSa.

\section{Conclusion}
In this paper, we propose a Relation and Sensitivity aware representation learning method (RaSa) for text-based person search, which contains two novel tasks, RA and SA, to learn powerful multi-modal representations. 
Given that the noise from the weak positive pairs tends to result in overfitting learning, the proposed RA utilizes an explicit detection between strong and weak positive pairs to highlight the difference, serving as a regularization of $p$-ITM that focuses on their consistency. 
Beyond learning transformation-insensitive representations, SA encourages the sensitivity to MLM-based token replacement.
Extensive experiments on multiple benchmarks demonstrate the effectiveness of RaSa.

\section*{Acknowledgments}
This work is supported by the National Science Foundation of China under Grant NSFC 62002252, and is also partially supported by the National Science Foundation of China under Grant NSFC 62106165.

\bibliographystyle{named}
\bibliography{rasa-arXiv}

\appendix
\section{Appendix}

\subsection{Datasets}
\label{datasets}
\textbf{CUHK-PEDES}~\cite{li2017person} is the most commonly-used dataset in text-based person search. It consists of $40,206$ images and $80,440$ texts from $13,003$ identities in total, which are split into $34,054$ images and $68,126$ texts from $11,003$ identities in the training set, $3,078$ images and $6,158$ texts from $1,000$ identities in the validation set, and $3,074$ images and $6,156$ texts from $1,000$ identities in the test set. The average length of all texts is $23$. 

\noindent\textbf{ICFG-PEDES}~\cite{ding2021semantically} is a recently published dataset, which contains $54,522$ images from $4,102$ identities in total. Each of the images is described by one text. The dataset is split into $34,674$ images from $3,102$ identities in the training set, and $19,848$ images from $1,000$ identities in the test set. On average, there are $37$ words for each text.

\noindent\textbf{RSTPReid}~\cite{zhu2021dssl} is also a newly released dataset to properly handle real scenarios. It contains $20,505$ images of $4,101$ identities. Each identity has $5$ corresponding images captured from different cameras. Each image is annotated with $2$ textual descriptions, and each description is no shorter than $23$ words. There are $3,701$/$200$/$200$ identities utilized for training/validation/testing, respectively.

\begin{table*}
\small
\renewcommand\arraystretch{1.2}
\setlength{\tabcolsep}{4.2pt}
\centering
\begin{tabular}{l|cccccc|cccccc} 
\hline
\multirow{3}{*}{Method} & \multicolumn{6}{c|}{Flickr30K (1K test set)}                & \multicolumn{6}{c}{COCO (5K test set)}                         \\
                        & \multicolumn{3}{c}{TR} & \multicolumn{3}{c|}{IR} & \multicolumn{3}{c}{TR} & \multicolumn{3}{c}{IR}  \\
                        & R@1   & R@5   & R@10   & R@1   & R@5   & R@10    & R@1   & R@5   & R@10   & R@1   & R@5   & R@10    \\ 
\hline
UNITER~\cite{chen2020uniter}    & 87.30 & 98.00 & 99.20 & 75.56 & 94.08 & 96.76   & 65.68 & 88.56 & 93.76  & 52.93 & 79.93 & 87.95   \\
COOKIE~\cite{wen2021cookie}     & 89.00 & 98.90 & 99.70 & 75.60 & 94.60 & 97.20   & 71.60 & 90.90 & 95.40  & 54.50 & 81.00 & 88.20   \\
Oscar~\cite{li2020oscar}        & - & - & - & - & - & -   & 73.50 & 92.20 & 96.00  & 57.50 & 82.80 & 89.80   \\
UNIMO~\cite{li2021unimo}        & 89.40 & 98.90 & 99.80 & 78.04 & 94.24 & 97.12   & - & - & -  & - & - & -   \\
ALIGN~\cite{jia2021scaling}     & 95.30 & \textbf{99.80} & \textbf{100.00} & 84.90 & 97.40 & 98.60   & 77.00 & 93.50 & 96.90  & 59.90 & 83.30 & 89.80   \\
BLIP~\cite{li2022blip}          & \textbf{97.40} & \textbf{99.80} & 99.90 & \textbf{87.60} & \textbf{97.70} & \textbf{99.00}   & \textbf{82.40} & \textbf{95.40} & \textbf{97.90}  & \textbf{65.10} & \textbf{86.30} & \textbf{91.80}   \\
\hline
ALBEF(backbone)~\cite{NEURIPS2021_50525975}                   & 95.50 & \textbf{99.80} & 99.90  & 85.44 & 97.34 & 98.70   & 77.26 & 94.02 & 97.04  & 60.31 & 84.22 & 90.51   \\
RaSa (Ours)                     & 96.00 & \textbf{99.80} & \textbf{100.00} & 85.90 & 97.54 & 98.72   & 77.44 & 94.12 & 97.18  & 61.00 & 84.49 & 90.83   \\
\hline
\end{tabular}
\caption{Results of coarse-grained retrieval on Flickr30K and COCO.}
\label{table5}
\end{table*}

\begin{table}
\small
\renewcommand\arraystretch{1.2}
\setlength{\tabcolsep}{1.5pt}
\centering
\begin{tabular}{l|cc|cc} 
\hline
\multirow{3}{*}{Method} & \multicolumn{2}{c|}{CUB}      & \multicolumn{2}{c}{Flowers}    \\
                                                    & TR            & IR            & TR            & IR  \\
                                                    & R@1           & AP@50         & R@1           & AP@50          \\ 
\hline
Bow~\cite{harris1954distributional}                 & 44.1          & 39.6          & 57.7          & 57.3           \\
Word2Vec~\cite{mikolov2013distributed}              & 38.6          & 33.5          & 54.2          & 52.1           \\
GMM+HGLMM~\cite{klein2015associating}               & 36.5          & 35.6          & 54.8          & 52.8           \\
Word CNN~\cite{reed2016learning}                    & 51.0          & 43.3          & 60.7          & 56.3           \\
Word CNN-RNN~\cite{reed2016learning}                & 56.8          & 48.7          & 65.6          & 59.6           \\
Triplet~\cite{li2017identity}                       & 52.5          & 52.4          & 64.3          & 64.9           \\
Latent Co-attention~\cite{li2017identity}           & 61.5          & 57.6          & 68.4          & 70.1           \\
CMPM/C~\cite{zhang2018deep}                         & 64.3          & 67.9          & 68.9          & 69.7           \\
TIMAM~\cite{sarafianos2019adversarial}              & 67.7          & 70.3          & 70.6          & 73.7           \\
GARN~\cite{jing2021learning}                        & 69.7          & 69.4          & 71.8          & 72.4           \\
DME~\cite{wang2021divide}                           & 69.4          & 71.8          & 72.4          & 74.6           \\
iVAD~\cite{wang2022improving}                       & 70.3          & 72.5          & 73.0          & 75.1           \\
\hline
\textbf{RaSa (Ours)}                         & \textbf{84.3}       & \textbf{84.5}     & \textbf{87.1}      & \textbf{84.3}               \\
\hline
\end{tabular}
\caption{Results of fine-grained retrieval on CUB and Flowers.}
\label{table6}
\end{table}

\subsection{Implementation Details}
\label{Implementation Details}
All experiments are conducted on $4$ NVIDIA $3090$ GPUs.
We train our model with $30$ epochs and a batch size of $52$. 
The AdamW optimizer~\cite{loshchilov2018decoupled} is adopted with a weight decay of $0.02$. 
The learning rate is initialized as $1e-4$ for the parameters of the classifiers in PRD and $m$-RTD, and $1e-5$ for the rest parameters of the model.
All images are resized to $384\times384$ and random horizontal flipping is employed for data augmentation.
The input texts are set with a maximum length of $50$ for all datasets.
The momentum coefficient in the momentum model is set as $m=0.995$.
The queue size $R$ is set as $65,536$ and the temperature $\tau$ is set as $0.07$ in CL.
The probability of inputting the weak positive pair is set as $p^{w}=0.1$ in RA, and the probability of masking the word in the text is set as $p^{m}=0.3$ in SA.
The hyper-parameters in the objective function are set as $\lambda_1=0.5$, $\lambda_2=0.5$, $\lambda_3=0.5$.  

\subsection{Extended Experiments}
\label{Extended Experiments}
We conduct extended experiments to verify the effectiveness of RaSa, including coarse-grained retrieval and fine-grained retrieval. 
Moreover, in order to verify the generalization ability of RaSa, we also conduct a series of domain generalization experiments, following LGUR~\cite{shao2022learning}.

\subsubsection{Coarse-grained Retrieval}
We consider two datasets for the coarse-grained retrieval task: Flickr30K~\cite{plummer2015flickr30k} and COCO~\cite{lin2014microsoft}.
Different from the text-based person search datasets with only one object (\emph{i.e.}, person) in the images and the fine-grained textual sentences, the images in Flickr30K and COCO contain various objects and the corresponding sentences usually present a coarse-grained description.
We follow the widely-used Karpathy split~\cite{karpathy2015deep} for both datasets.
The images in Flickr30K are split into $29$K/$1$K/$1$K and the images in COCO are split into $113$K/$5$K/$5$K for training/validation/testing, respectively.
Each image in both two datasets is annotated by five sentences.


It should be noted that each image together with the paired texts is a unique class in the two datasets, as a result of which there is no intra-class variation in the images and all of positive image-text pairs belong to the strong positive type.
Therefore, the proposed RA, which aims at differentiating between strong and weak positive pairs, no longer applies to the experiments on Flickr30K and COCO.
\textbf{We only perform SA and use the vanilla ITM for the experiments.}



As shown in Table~\ref{table5}, RaSa achieves a comparable performance compared with existing methods. 
Particularly, compared with the backbone model ALBEF~\cite{NEURIPS2021_50525975}\footnote{We report the results reproduced with the released code of ALBEF, where the batch size is set as same as the introduction in Appendix~\ref{Implementation Details} for a fair comparison.}, RaSa with only SA still brings consistent improvement in terms of all metrics.
We argue that SA constructs a non-trivial pretext task to explicitly endow the model with the ability to perceive the sensitive transformation, which significantly facilitates the representation learning and eventually gains a better performance.

\begin{table}
\small
\centering
\renewcommand\arraystretch{1.2}
\setlength{\tabcolsep}{5pt}
\begin{tabular}{c|l|ccc}
\hline
                                                        & Method                            & R@1       & R@5       & R@10       \\
\hline
\multirow{6}{*}{\rotatebox{90}{C $\rightarrow$ I}}      & Dual Path \cite{zheng2020dual}    & 15.41     & 29.80      & 38.19           \\
                                                        & MIA \cite{niu2020improving}       & 19.35     & 36.78      & 46.42          \\
                                                        & SCAN \cite{lee2018stacked}        & 21.27     & 39.26      & 48.83            \\
                                                        & SSAN \cite{ding2021semantically}  & 29.24     & 49.00      & 58.53        \\
                                                        & LGUR \cite{shao2022learning}      & 34.25     & 52.58      & 60.85        \\
                                                        \cline{2-5}
                                                        & \textbf{RaSa (Ours)} & \textbf{50.59} & \textbf{67.46} & \textbf{74.09}  \\
\hline\hline
\multirow{6}{*}{\rotatebox{90}{I $\rightarrow$ C}}      & Dual Path \cite{zheng2020dual}    & 7.63     & 17.14      & 23.52       \\
                                                        & MIA \cite{niu2020improving}       & 10.93     & 23.77      & 32.39       \\
                                                        & SCAN \cite{lee2018stacked}        & 13.63     & 28.61      & 37.05       \\
                                                        & SSAN \cite{ding2021semantically}  & 21.07     & 38.94      & 48.54       \\
                                                        & LGUR \cite{shao2022learning}      & 25.44     & 44.48      & 54.39       \\
                                                        \cline{2-5}
                                                        & \textbf{RaSa (Ours)} & \textbf{50.70} & \textbf{72.40} & \textbf{79.58}  \\
\hline
\end{tabular}
\caption{Comparison with other methods on domain generalization task. We adopt CUHK-PEDES (denoted as C) and ICFG-PEDES (represented as I) as the source domain and the target domain in turn.}
\label{table7}
\end{table}

\begin{table}[t]
\small
\centering
\tabcolsep=2pt
\renewcommand\arraystretch{1.1}
\begin{tabular}{c|l|cccc}
\hline
                                               & Method    & R@1  & R@5  & R@10   & mAP \\
\hline
\multirow{11}{*}{\rotatebox{90}{w/o VLP}}      & GNA-RNN \cite{li2017person}      & 19.05     & -      & 53.64      & -      \\
                                               & Dual Path \cite{zheng2020dual}   & 44.40     & 66.26      & 75.07      & -      \\
                                               & CMPM/C \cite{zhang2018deep}      & 49.37     & 71.69      & 79.27      & -      \\
                                               & ViTAA \cite{wang2020vitaa}       & 55.97     & 75.84      & 83.52      & -      \\
                                               & DSSL \cite{zhu2021dssl}          & 59.98     & 80.41      & 87.56      & -      \\
                                               & MGEL \cite{wang2021text}         & 60.27     & 80.01      & 86.74      & -      \\
                                               & ACSA \cite{ji2022asymmetric}     & 63.56     & 81.40      & 87.70      & -  \\
                                               & SAF \cite{li2022learning}        & 64.13     & 82.62      & 88.40      & 58.61  \\
                                               & TIPCB \cite{chen2022tipcb}       & 64.26     & 83.19      & 89.10      & -      \\
                                               & CAIBC \cite{wang2022caibc}       & 64.43     & 82.87      & 88.37      & -      \\
                                               & $\rm C_2A_2$ \cite{niu2022cross} & 64.82     & 83.54      & 89.77      & -  \\
                                               & LGUR \cite{shao2022learning}     & 65.25     & 83.12      & 89.00      & -  \\
\hline\hline
\multirow{7}{*}{\rotatebox{90}{w/ VLP}}        & PSLD \cite{han2021textreid}      & 64.08     & 81.73      & 88.19      & 60.08  \\
                                               & IVT \cite{shu2022see}            & 65.59     & 83.11      & 89.21      & -  \\
                                               & CFine \cite{yan2022clip}         & 69.57     & 85.93      & 91.15      & -  \\
\cline{2-6}
                                               & CLIP~\cite{radford2021learning}     & 43.05 & 66.41 & 76.36 & 38.91  \\
                                               & RaSa$_\text{CLIP}$                   & 57.60 & 78.09 & 84.91 & 55.52  \\
                                               & TCL~\cite{yang2022vision}           & 57.60 & 77.14 & 84.39 & 53.64  \\
                                               & \textbf{RaSa$_\text{TCL}$}   & \textbf{73.23} & \textbf{89.20} & \textbf{93.32} & \textbf{66.43}  \\
\hline
\end{tabular}
\caption{Comparison with other methods on CUHK-PEDES. RaSa$_\text{CLIP}$ adopts CLIP as the backbone, while RaSa$_\text{TCL}$ uses TCL as the backbone.}
\label{table8}
\end{table}

\subsubsection{Fine-grained Retrieval}
Apart from the fine-grained retrieval task of text-based person search, we furthermore evaluate RaSa on other fine-grained datasets: CUB~\cite{reed2016learning} and Flowers~\cite{reed2016learning}.
CUB contains $11,788$ bird images from $200$ different categories, and each image is annotated with $10$ sentences. The dataset is split into $100$/$50$/$50$ categories for training/validation/testing, respectively.
Flowers consists of $8,189$ flower images from $102$ categories, and each image has $10$ descriptions. There are $62$, $20$ and $20$ categories utilized for training, validation and testing, respectively.

Following common settings~\cite{reed2016learning,sarafianos2019adversarial}, we take random cropping and horizontal flipping as the data augmentation, and the maximum length of the input texts is set as $30$.
Other settings are kept as same as the introduction in Appendix~\ref{Implementation Details}.
Therein, we use AP@50 metric for the evaluation of text-to-image retrieval and R@1 for image-to-text retrieval, where AP@50 reflects the average matching percentage of top-50 retrieved images of all test text classes.
During inference, existing methods usually compute the metrics according to the similarity between the image embedding and the average of the corresponding text embeddings.
However, since RaSa is a one-stream model and its final output is the multi-modal embedding rather than the text embedding, we compute the metrics by averaging the multi-modal embeddings of the same identity.


From Table~\ref{table6}, RaSa outperforms all existing state-of-the-art methods by a large margin.
Specifically, compared with iVAD~\cite{wang2022improving}, the performance of RaSa has $14.0$\% and $12.0$\% improvements on CUB and $14.1$\% and $9.2$\% boosts on Flowers in terms of R@1 and AP@50, respectively.
It is worth noting that existing methods ignore the noise interference caused by the weak positive pairs and model all positive relations without distinction.
Inevitably, they are vulnerable to overfitting learning.
On the contrary, RaSa utilizes RA to explicitly distinguish different types of positive relation and SA to learn more robust representations.
As a result, it achieves a decent performance.

\subsubsection{Domain Generalization}
We conduct a series of domain generalization experiments to investigate the generalization ability of RaSa.
Specifically, we use the model trained on the source domain to evaluate the performance on the target domain, where CUHK-PEDES and ICFG-PEDES are adopted as the source domain and the target domain in turn.

As shown in Table~\ref{table7}, RaSa outperforms other methods by a large margin. 
We conjecture that there exist two factors bringing such a significant improvement.
(1) Other methods are inclined to overfitting learning since they neglect the noise interference from the weak positive pairs, while RaSa substantially alleviates the effect of the noise and is able to learn more robust representations.
(2) The parameters of RaSa are initialized from the VLP models which contain abundant multi-modal knowledge and eventually facilitate representation learning.
Overall, the results on the domain generalization task effectively demonstrate the powerful generalization ability of RaSa.

\subsection{Backbones and Experiments}
\label{backbones}
Apart from ALBEF~\cite{NEURIPS2021_50525975}, we also apply RaSa on other backbones: TCL~\cite{yang2022vision} and CLIP~\cite{radford2021learning}.

\textbf{TCL} has a similar architecture with ALBEF and is pretrained on $4$M image-text pairs.
\textbf{CLIP} is pretrained on $400$M image-text pairs and is comprised of two unimodal encoders to individually process the images and texts.
However, the proposed RaSa works on the multi-modal features from the cross-modal encoder.
Therefore, we additionally append a one-layer transformer block on the outputs of CLIP as the cross-modal encoder when adopting CLIP as the backbone. 

As shown in Table~\ref{table8}, no matter whether TCL or CLIP is adopted as the backbone, RaSa always brings consistent improvements in terms of all metrics.
Meanwhile, a stronger backbone can lead to a better performance.
For example, in terms of R@1, RaSa$_\text{TCL}$ achieves the best performance with $73.23$\%,
while RaSa$_\text{CLIP}$ achieves a modest performance of $57.60$\%.
We conjecture that (1) the lack of cross-modal deep fusion in the backbone CLIP makes the model difficult to capture fine-grained details, which tends to have a negative impact to the performance, and (2) the parameters of the one-layer transformer block are randomly initialized, rendering the model inclined to be trapped in the local minimum.

\begin{figure}[t]
\centering
\includegraphics[width=1\linewidth, height=1.4\linewidth]{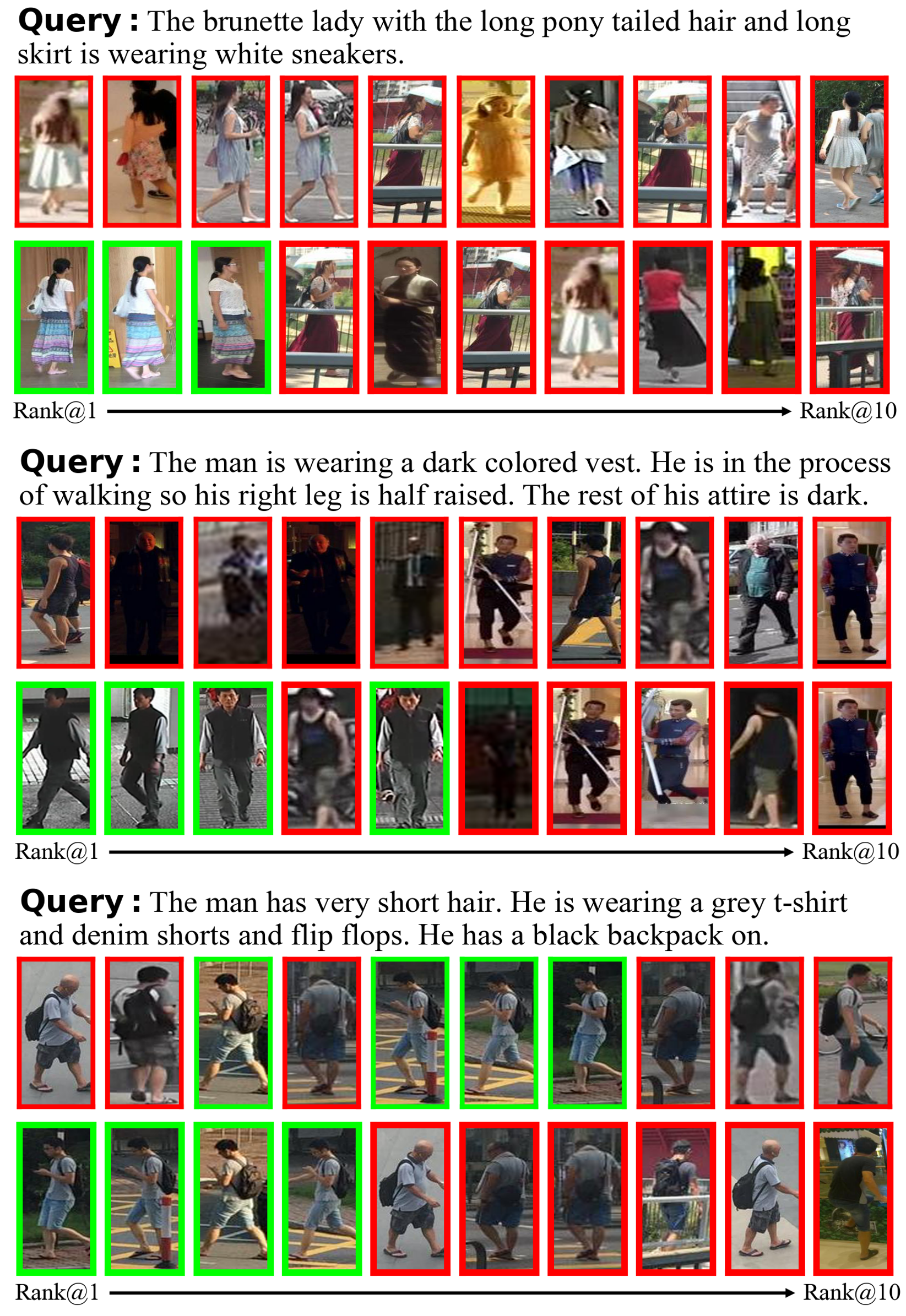} 
\caption{Visualization of top-10 retrieval results on CUHK-PEDES.
The first row in each example presents the retrieval results from the backbone ALBEF, and the second row shows the results from RaSa.
Correct/Incorrect images are marked by \textcolor{green}{green}/\textcolor{red}{red} rectangles.}
\label{fig5}
\end{figure}

\subsection{Visualization}
\label{visualization}
We exhibit three top-10 retrieval examples of the backbone ALBEF~\cite{NEURIPS2021_50525975} and RaSa in Figure~\ref{fig5}, where the first row and the second row in each example present the retrieval results from ALBEF and RaSa, respectively.
It can be seen that RaSa can retrieve the corresponding pedestrian images for a query text more accurately. 
This is mainly due to the alleviation of the noise interference in RA and the powerful sensitivity-aware learning strategy in SA.
The visualization vividly demonstrates the effectiveness of RaSa.

\end{document}